\documentclass[11pt]{article}
   
\usepackage{authblk}
\usepackage{ACL2023}
\usepackage{CJKutf8}
\usepackage{times}
\usepackage{latexsym}
\usepackage{amsmath}
\usepackage{booktabs}
\usepackage{multirow}
\usepackage{graphicx}
\usepackage{pdfpages}
\usepackage[export]{adjustbox}
\usepackage{enumitem}
\usepackage[T1]{fontenc}
\usepackage[utf8]{inputenc}

\usepackage{microtype}

\usepackage{inconsolata}

\usepackage{etoolbox}
\makeatletter
\patchcmd{\maketitle}
 {\def\@makefnmark}
 {\def\@makefnmark{}\def\useless@macro}
 {}{}
\makeatother

\title{Automatically Suggesting Diverse Example Sentences for L2 Japanese Learners Using Pre-Trained Language Models}

\author[1*\thanks{\textsuperscript{*}Research conducted during internship at NII, Japan.}]{Enrico Benedetti}
\author[2]{Akiko Aizawa}
\author[2,3]{Florian Boudin}
\affil[1]{University of Bologna, Italy}
\affil[2]{National Institute of Informatics, Japan}
\affil[3]{JFLI, CNRS, Nantes University, France}
\affil[ ]{\texttt{enrico.benedetti5@studio.unibo.it\\aizawa@nii.ac.jp}}
\affil[ ]{\texttt{florian.boudin@univ-nantes.fr}}

\begin{document}

\maketitle

\begin{abstract}
Providing example sentences that are diverse and aligned with learners' proficiency levels is essential for fostering effective language acquisition.
This study examines the use of Pre-trained Language Models (PLMs) to produce example sentences targeting L2 Japanese learners.
We utilize PLMs in two ways: as quality scoring components in a retrieval system that draws from a newly curated corpus of Japanese sentences, and as direct sentence generators using zero-shot learning.
We evaluate the quality of sentences by considering multiple aspects such as difficulty, diversity, and naturalness, with a panel of raters consisting of learners of Japanese, native speakers -- and GPT-4.
Our findings suggest that there is inherent disagreement among participants on the ratings of sentence qualities, except for difficulty. 
Despite that, the retrieval approach was preferred by all evaluators, especially for beginner and advanced target proficiency, while the generative approaches received lower scores on average.
Even so, our experiments highlight the potential for using PLMs to enhance the adaptability of sentence suggestion systems and therefore improve the language learning journey.

\end{abstract}

\section{Introduction}

\begin{figure}[ht!]
    \centering
    \smallskip\noindent
    \includegraphics[max size={0.50\textwidth}{0.4\textheight}]{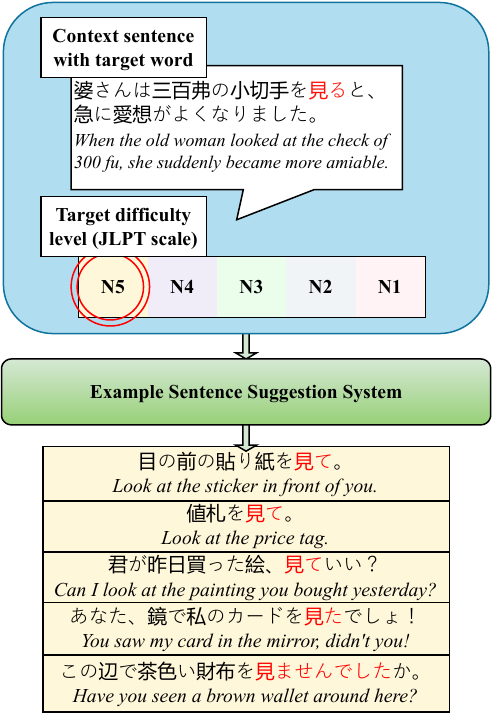}
    \caption{Task overview. Given a word in context and a difficulty level, the system will suggest diverse and level-appropriate examples. In this instance, the target is \textit{miru}, to see.}
    \label{fig:overview_task}
\end{figure}

%Briefly introduce the problem and its relevance in NLP.
% talk about linguistic stuff and why it is important for learners

The term second language acquisition (or L2 acquisition) refers to the process of learning a second language by those who already know a first one. While children have a natural predisposition for acquiring languages, the degree of success among L2 learners varies greatly, as it is usually harder in adult life, requiring a combination of conscious effort, motivation, support from teachers and adequate materials~\cite{fromkin2013introduction}.

Online dictionaries are usually the first resource towards which learners turn to in order to understand an unknown word or expression via definitions and example sentences.
%However, producing high-quality learning material requires effort and expert knowledge, which is why there is much research on automatic example selection and generation directed to help professionals such as lexicographers or teachers and non-experts such as language learners~\cite{kilgarriff_gdex_2008, borthwick_icalls_2017, volodina_selection_icall}.
However, producing high-quality learning material requires effort and expert knowledge. Because of that, researchers have explored automated techniques for selecting and generating examples to aid professionals like lexicographers or teachers, as well as non-experts like language learners~\cite{kilgarriff_gdex_2008, borthwick_icalls_2017, volodina_selection_icall}.

%Recently, websites of linguistic resources, chat bots and language learning applications are also widely used for this aim. %maybe remove last part?

% There are also methods for remembering vocabulary efficiently based on Spaced Repetition (SR), which is the concept that studying and testing oneself at different time intervals over a long period is usually better than studying by cramming. For example, Anki is one of the most well-known open source Spaced Repetition System (SRS).
% %also other paid apps use a similar concept
% Learners can study vocabulary thorugh SR by using flashcards with a sentence or target word on one side, and the definition on the other side. Sentence Mining is the process of obtaining that kind of study material.

% Methods for obtaining good example sentences automatically could help reduce the effort in creating a personalized study collection.

% research question
%The main motivation for this work is to investigate whether pre-trained Language Models (PLMs) %can help language learners encounter a broad range of uses for the target words they are interested in, by proposing understandable and diverse sentences.
Pre-trained Language Models (PLMs) have been shown to be effective for many NLP tasks~\cite{WANG202351}.
The main motivation for this work is to investigate whether PLMs can be leveraged to propose sentences that are understandable and diverse to help L2 learners be exposed to a broad range of uses for the target words they are interested in (e.g. an unknown word encountered while reading), since examples contribute to improving vocabulary knowledge~\cite{baicheng2009}.

In this study, we focus on Japanese, as an increasing number of people are interested in achieving a certain level of proficiency, be it for study, work, culture or other reasons ~\cite{jlpt_nodate}.
While there is substantial work on obtaining high-quality text from corpora or generative models, as discussed in Section~\ref{sec:rel_work}, to the best of our knowledge, there are few studies simultaneously addressing the Japanese example sentence suggestion task, and developments in Natural Language Processing (NLP) such as the emergence of PLMs. The existing work mostly focuses on functional expressions~\cite{liu_automatic_2018, liu_sentence_2018, liu_simplification_2016, shortt_synthesizing_2021} or exercises~\cite{educsci13121203}.

% oldv
% there is much work on obtaining high-quality text from corpora or from generative models (see Section \ref{sec:rel_work}).
% However, to the best of our knowledge there are few studies that address the task of example sentences suggestion in Japanese considering the latest developments in Natural Language Processing (NLP) and the impact of PLMs. The existing work mostly focuses on functional expressions~\cite{liu_automatic_2018, liu_sentence_2018, liu_simplification_2016, shortt_synthesizing_2021} or exercises~\cite{educsci13121203}.

Our contributions are summarized as follows:
\begin{enumerate}[nosep]
    \item We develop a retrieval-based approach to select example sentences from a corpus, by combining different PLM modules and NLP techniques for scoring sentence quality according to four criteria: difficulty, sense similarity, syntactic and lexical diversity.
    
    \item We build WJTSentDiL, a corpus of sentences from different web sources, annotated with Japanese Language Proficiency Test\footnote{More details on the \href{https://www.jlpt.jp/e/index.html}{JLPT website} and Section \ref{sec:difficulty}.} (JLPT) labels.

    \item We evaluate the quality of selected example sentences for specific target words by comparing the retrieval approach to two generative PLM baselines, employing native speakers and learners, alongside GPT-4~\cite{openai2023gpt4}. We present the insights obtained from the investigation.
\end{enumerate}
The main repository for this work can be found here: \href{https://github.com/EnricoBenedetti/NihongoExamplePLM}{NihongoExamplePLM}. 

%\item We compare the retrieval approach and two generative PLM baselines by employing volunteer native speakers and learners, as well as the text generation model GPT-4-turbo~\cite{openai2023gpt4} to evaluate sentence quality for a selected group of target words, and present the insights obtained from the investigation.
% they will be explained more in detail in the task section
%(3) We propose a simple BERT-based text classifier for estimating sentence difficulty on the JLPT scale.
%(4) We apply a pre-existing method to fine-tune a model for contextual word embeddings. %MirrorWiC by \citet{liu_mirrorwic_2021} 
%(5) We carry out a human study with native speakers and learners comparing the retrieval approach and two PLM baselines in the zero-shot setting.
%(6) We investigate whether a very capable text generation model, GPT-4-turbo \cite{openai2023gpt4}, can be reliable in evaluating sentence quality in this task.
%(6) We also employ a text generation model, GPT-4-turbo~\cite{openai2023gpt4} in evaluating sentence quality in this task and present the insights obtained from the investigation.
% (2) We propose a sentence corpus annotated with JLPT difficulty level. % Insert dataset name here
% (3) We propose a simple model for sentence difficulty estimation based on the JLPT. (4) We finetune with a pre-existing method a model for contextual word embeddings.
% (5) We combine together the previous resources into a baseline approach for example sentence suggestion.

\section{Related Work}
\label{sec:rel_work}
In the following we discuss the related work, namely retrieving and generating example sentences, and estimating sentence difficulty.

%\subsection{Example selection}
\paragraph{Example selection}
Similarly to~\citet{tolmachev-kurohashi-2017-automatic}, we seek to provide high-quality and diverse example Japanese sentences. They propose a thorough retrieval approach based on quality and diversity scoring using a Determinantal Point Process, and carry out an evaluation with L2 learners and a teacher.
Our work differs from theirs in that we focus on selecting sentences for sense similarity given a target word in context, instead of many possible senses for a word in isolation. Furthermore, we evaluate more aspects of the systems, in particular their capacity to adapt to learner proficiency levels. We also employ a language model in the evaluation.

%We also perform a small-scale human evaluation with L2 learners and native speakers, analyzing comments and requiring many judgments in

%Our evaluation is small-scale, involving a limited number of L2 learners and native speakers, but we obtain ratings on multiple sentence qualities in addition to preferences, and consider responses of an advanced Large Language Model (LLM) GPT-4-turbo.

Many other works deal with the task of example sentence selection from a corpus, focusing on dictionary examples for English, Japanese and Swedish~\cite{kilgarriff_gdex_2008, de-melo-weikum-2009-extracting, hazelbeck_corpus-based_2009, pilan_automatic_2013}. Additionally, \citet{Shinnou2008DivisionOE}, \citet{kathuria_word_2012} and \citet{cheng_languagenet_2018} leverage parallel corpora to extract disambiguated sentences, while we limit our experiments to the monolingual setting.

%\subsection{Example generation}
\paragraph{Example generation}
There is a lot of research on controllable text generation approaches~\cite{survey_generation}.
Possible generation targets are definitions for a given term~\cite{zhang-etal-2023-assisting, Gardner2022DefinitionML}, as well as example sentences. When it comes to example generation, researchers have shown that generated sentences can improve performance in Word Sense Disambiguation tasks in a supervised~\cite{barba_exemplification_2021} or unsupervised way~\cite{he_controllable_2022}.
Focusing on L2 learners, \citet{harvill_one-shot_2023} consider lexical complexity and sentence length to generate example sentences of controllable difficulty.
% when talking about WSD
%In our case, we decide to not depend on fixed sense inventories, although using dictionary definitions could prove beneficial to learners in future research.
In our case, we opt not to rely on fixed sense inventories, primarily due to the scarcity of available sense-tagged corpora. However, we believe that assigning dictionary definitions to words could prove beneficial to learners.% in future research.

% old rewriting? 
%In our case we resort to not depending on fixed sense inventories due to the lack of public and freely available sense-tagged corpora for Japanese, but we believe that using dictionary definitions could prove beneficial to learners in future research.

% reply <---
%The main reason is that given our resources there was no convenient way of linking (for example) dictionary definitions to words.% (freely available, there's only the EDICT dictionary project - to my knowledge).
%To do that one would require a model, or data to train one (sense-tagged corpora), which I could not find. There was the corpus of balanced Japanese but it was not free. Therefore, we worked on monolingual, open data and unsupervised / weakly supervised methods only.
%Also it was to keep it simple, and the work I based my research on did not use sense inventories (they did not explain why but I think it could be because of similar reasons..)

%\subsection{Sentence difficulty estimation} 
\paragraph{Sentence difficulty estimation} 
% intro on general sentence difficulty estimation?
Determining the level of difficulty of text is a key challenge in educational NLP, as vocabulary and grammatical structure interact in a complex way~\cite{text_diff_collins}.
To estimate the difficulty of Japanese sentences, \citet{jlpt_nodate} show that a BERT-based classifier~\cite{devlin-etal-2019-bert} trained on labeled examples can achieve good performance, surpassing existing readability metrics\footnote{\href{https://jreadability.net/sys/en}{https://jreadability.net/sys/en}} and approaches based on word frequencies.
\citet{liu-matsumoto-2017-sentence} focus on estimating Japanese text difficulty for learners with pre-existing knowledge of Chinese characters. In that case, the main source of difficulty is not vocabulary, but grammar and functional expressions.
In our work, due to lacking training data from official JLPT material, we train a similar classifier to \citet{jlpt_nodate}.
% by collecting sentences from publicly available language learning websites, and test on official JLPT exam data freely available online.

\section{Task: Example Sentence Suggestion}

% describe the task, in detail
We define the L2 contextualized example suggestion task as: %$$M(w, s_0, d) = [s_i \ \text{for} \ i : 1, K ]$$
\begin{equation}
\label{eq:task}
    M(w, s_0, d) = \{s_1, s_2, \ldots, s_i, \ldots, s_K\}
\end{equation}
Given a target word $w$, a context sentence $s_0$ and a target difficulty level $d$, we want to obtain a list of $K$ good example sentences from a model $M$.

To expand more on what makes a good example,
\citet{kilgarriff_gdex_2008} suggest that such examples should represent typical usage, be informative and understandable to learners.
Building upon the discussion presented by \citet{Arseny_Tolmachev2022}, we aim to obtain multiple examples with diverse syntactic patterns since learners preferred them.

\section{Methodology}

% Present your approach using retrieval and generative models.
% Summarize data sources (Wikipedia, Tatoeba, jpWaC).
% Briefly explain the LLM and ChatGPT models used.
%In the following, we present our retrieval approach for example sentence selection. We wish to investigate its performance against PLM baselines. 

\subsection{Retrieval method}
\label{sec:retr}
We design a retrieval model that, given a query, will select candidate sentences containing a target word from a corpus and present them to the learner (for more details on the corpus, see Section \ref{sec:corpus}).
Candidate sentences are ranked by how closely they match the target difficulty level and the semantic similarity of the target word in both the suggested and context sentences.
Finally, the model selects a subset of sentences considering the total diversity of the list.
In summary, we devise a model to quantify for a sentence $s_i$:
	\begin{enumerate}[nosep]
 		\item how adequate $s_i$ is with respect to the target difficulty level $d$ (Sec. \ref{sec:difficulty}).
		\item if $s_i$ contains the target word $w$ and it is used in the same sense as the target word of the context sentence (Sec. \ref{sec:sense}).
		\item the diversity of $\{s_0, s_1, s_2, \ldots, s_i, \ldots, s_K\}$ on vocabulary and syntax (Sec. \ref{sec:diversity}).
	\end{enumerate}

\subsubsection{Quality: difficulty}
\label{sec:difficulty}
 The Japanese Language Proficiency Test (JLPT) has a proficiency scale similar to the Common European Framework of Reference for Languages (CEFR). The JLPT levels are, from easier to harder: N5, N4, N3, N2 and N1.
 Our classifier will therefore assign a JLPT level $d_i$ to input sentences.
Then, it will be mapped to a difficulty score between $1$ and $0$.
We formulate this score as 
\begin{equation}
    max\big(0, 1 - \text{penalty}_\text{diff} * (d - d_i)\big)
\end{equation}
where $d$ and $d_i$ are the target difficulty level and difficulty label of the sentence $i$.
We manually set the coefficient $\text{penalty}_\text{diff}$ to $0.2$.
We increase the coefficient to $0.4$ on sentences deemed harder than the target level because L2 learners might benefit more from easier sentences in case of discrepancies.

\subsubsection{Quality: sense similarity}
\label{sec:sense}

%\citet{anderson-camacho-collados-2022-assessing} and
\citet{pilehvar-camacho-collados-2019-wic} propose Words in Context (WiC), a different declination of Word Sense Disambiguation. WiC is a binary classification task: given a target word and two contexts, the model has to predict whether the word is used with the same meaning.
Since we also tackle this problem in our case, we turn to MirrorWiC, an unsupervised fine-tuning method for contextualized word sense embeddings \cite{liu_mirrorwic_2021}. 
We fine-tune a PLM with MirrorWiC and use the resulting model to extract a vector representation for the target words in context.
Then, we assign a sense similarity score based on cosine similarity between $s_0$, the context sentence, and $s_i$.

\subsubsection{Diversity: syntactical and lexical}

\label{sec:diversity}
Inspired by the way \citet{tolmachev-kurohashi-2017-automatic} measured syntax diversity, we opt for a simpler approach, supported by other works on syntax similarity~\cite{chen2023fastkassim, kanagawa2016syntax}.

We compute dependency trees of two sentences and partially generalize their labels, then apply a Label-based Tree Kernel Similarity method, FastKASSIM, to obtain a diversity score \cite{chen2023fastkassim, moschitti2006making, boghrati2018conversation}.
More in detail, we compute the parse trees and the number of shared subtrees of a pair of sentences. The latter is normalized with the square root of the product of the number of subtrees for each sentence \cite{chen2023fastkassim}.
For the syntactic diversity of a list of sentences, we take the average of pairwise scores.
% \subsubsection{Syntax diversity}
% To obtain a diversity score, we employ the methodology described in Section \ref{sec:syntaxmet}.
%We use SpaCy and the parser Ginza to compute a dependency parse trees for sentences, and substitute its labels with Part of Speech (PoS) and dependency labels. %(show example?) there may not be enough space?

For lexical diversity, we simply compute the average percentage of unique 1-2-3-4-grams in a sentence list.
%also considering the context sentence.
%we computed the average percentage of unique n-grams across four different sizes: unigrams (1-grams), bigrams (2-grams), trigrams (3-grams), and four-grams (4-grams).

Finally, we obtain a combined diversity score by equally weighting the lexical and syntax scores.

\subsubsection{Ranking and Greedy Selection}

%We build a set of $K$ final sentences using a greedy algorithm by including the context sentence as the first ``dummy'' element, then adding to the list the sentence (among the top scoring candidate sentences in terms of difficulty and sense) which also has the highest diversity score, obtained by equally weighting lexical and syntax scores.

% rephrase the above
 
As the number of candidates can be very high, we greedily select $K$ final sentences.
First, we sort the candidate sentences in terms of difficulty and sense scores, having equal weights as we considered the qualities equally important for this experiment.
Then, within a window, we iteratively add the sentence which achieves the highest diversity score, until the list is complete.
We set a window of only $50$ candidates in the preliminary experiments. Otherwise, queries would take a long time due to having to re-compute similarity scores for every partial list.

\subsection{PLM generation method} 
Considering the PLM baselines, we prompt them with the query, expressed in English.
We share the prompt used in Appendix \ref{sec:prompts}.
As initial experiments revealed that complying with the query in zero-shot manner was quite difficult, we prompt the PLMs multiple times, concatenate the outputs and exclude duplicates and sentences without the target word, until we get the required number of sentences. In the majority of cases, twice was enough.
We set the generation temperature parameter to $1.0$ for all PLMs; additionally, for LLM-jp, we add a repetition penalty of $5.0$.

\section{Experimental setup}
\subsection{Dataset: WJTSentDiL Corpus}
\label{sec:corpus}

We present WJTSentDiL,\footnote{The corpus is available on \href{https://huggingface.co/datasets/bennexx/WJTSentDiL}{HuggingFace}.} a corpus of \textbf{W}ikipedia, \textbf{J}pWaC and \textbf{T}atoeba \textbf{Sent}ences with \textbf{Di}fficulty \textbf{L}evel. It is built by merging together three public corpora (described below) and performing additional filtering to remove spurious sentences. Additionally, our difficulty classifier adds JLPT levels to each sentence.
%
% item list of the sources
\begin{itemize}[nosep]
    \item \href{https://tatoeba.org/en/}{Tatoeba} is a platform where users can share sentences and translations. 
We select only Japanese sentences and fix errors where entries are made from multiple sentences.

\item JpWaC \cite{Sangawa2010AutomatedCO} is a curated corpus of sentences automatically collected from Japanese web domains.
We include subsets L0 to L4 of the corpus.

\item Wikipedia is a free online encyclopedia.  We process raw article text from the Japanese part of the website, more specifically the \href{https://dumps.wikimedia.org/jawiki/20231201/jawiki-20231201-pages-articles-multistream.xml.bz2}{``jawiki dump''} from December 2023.
\end{itemize}

We use spaCy\footnote{\href{https://github.com/explosion/spaCy}{Repository for spaCy}, version 3.7.2} and Ginza\footnote{\href{https://github.com/megagonlabs/ginza}{Repository for ginza}, version 5.1.3, `ja-ginza' model.} to split raw text into sentences, tokenize them, and assign part-of-speech (POS) tags.
To keep well-formed sentences, we apply filters following heuristics similar to \citet{kilgarriff_gdex_2008} and \citet{Sangawa2010AutomatedCO}.
Namely, we keep sentences that:
\begin{itemize}[nosep]
    \item have a length between $5$ and $50$ tokens.
    \item have less than $20\%$ punctuation or numerals.
    \item do not contain tokens from the Latin, Cyrillic and Arabic scripts.
    \item end in a predicate and punctuation, or particles such as \begin{CJK}{UTF8}{min}よ, ね\end{CJK}.
    \item are not duplicates.
\end{itemize}
Wikipedia sentences are what makes up most of the corpus. They are on average longer and contain more \textit{kanji}, Chinese characters, compared to the other sources.
We show statistics in Table \ref{tab:corpus}.
% show table of statistics
% \begin{table*}[h]
% \centering
%             \begin{tabular}{lrrrrrrrrr}
%                 \toprule
%                 Corpus    & Sentences   & avg tokens  & max & min  & std   & avg kanji ratio & $\%$ of total \\
%                 \midrule
%                 jpWaC     & 152,751   & 13.01 & 31  & 2   & 5.69      & 0.27                       & 0.01 \\
%                 Tatoeba   & 245,793   & 11.07 & 69  & 2   & 4.59     & 0.27                       & 0.02 \\
%                 Wikipedia & 12,306,416 & 26.39 & 52  & 4   & 10.60   & 0.37                      & 0.97 \\
%                 \bottomrule
%                 WJTSentDiL     & 12,704,960 & 25.93 & 69  & 2   & 10.78    & 0.36                     & 1.00 \\
%                 \bottomrule
%             \end{tabular}
%             \label{tab:corpus}
%             \caption{Statistics of WJTSentDiL by source. 
%             ``Sentences'' is the count of sentences, while ``avg'', ``max'', ``min'', ``std'' refer to the average, maximum, minimum and standard deviation of the token counts of the sentences, respectively.
%             The column ``avg kanji ratio'' indicates the proportion between kanji (Chinese characters) and other characters.}
% \end{table*}

% shorter version
\begin{table}[ht]
\centering
\resizebox{\linewidth}{!}{%
    \begin{tabular}{l|rrr r}
    \toprule
    \textbf{Corpus}   & \textbf{Sentences}   & \textbf{Tokens}  & \textbf{Kanji} (\%) & \textbf{Ratio} (\%)  \\
    \midrule
    JpWaC     & 152\,751   & 13.01     & 27.31    &   1.2         \\
    Tatoeba   & 245\,793   & 11.07      & 26.75    &  1.9           \\
    Wikipedia & 12\,306\,416 & 26.39     & 36.67   & 96.9             \\
    \midrule
    WJTSentDiL     & 12\,704\,960 & 25.93    & 36.35  & 100         \\
    \bottomrule
\end{tabular}
}
\caption{Statistics of WJTSentDiL by source.
    ``Tokens'' is the average token count.%, from Ginza's tokenizer.
        ``Kanji'' reports the proportion between Chinese characters and the rest.}
\label{tab:corpus}
\end{table}

\subsection{Retrieval method details}

\subsubsection{Inverted index}
%\paragraph{Inverted index}
The retrieval model uses an inverted index, mapping words to sentences they appear in. The keys are lemmas or ``dictionary forms'' of words and compound nouns.
The candidate sentences are retrieved using the index by lemmatizing the target word. For example, the target word ``\begin{CJK}{UTF8}{min}たべた\end{CJK}'' (past form of \textit{to eat}) is lemmatized as ``\begin{CJK}{UTF8}{min}食べる+た\end{CJK}'' (\textit{to eat} $+$ past tense auxiliary verb).

\subsubsection{Difficulty classifier}
\label{sec:diff_theory}
%\paragraph{Difficulty classifier}
%  Then, we use the predictions from the JLPT level classifier to assign a difficulty label to the sentence.
The JLPT difficulty classifier is a BERT model pre-trained on texts in the Japanese language,\footnote{\href{https://huggingface.co/tohoku-nlp/bert-base-japanese-v3}{tohoku-nlp/bert-base-japanese-v3}} that we fine-tuned on 5,000 sentences from Japanese language learning websites.\footnote{\href{https://nihongokyoshi-net.com/}{nihongokyoshi-net.com}, \href{https://jlptsensei.com/}{jlptsensei.com}.
Due to license limitations, we can not share the sentences, but the model is available on \href{https://huggingface.co/bennexx/cl-tohoku-bert-base-japanese-v3-jlpt-classifier}{HuggingFace}.}
Their labels are assigned based on HTML metadata specific to each website.
% it was done by other researchers so...
%\footnote{We can provide the test dataset and model weights, but not the training data because of the websites' copyright policy.} 
 For more details on the training and evaluation of the classifier, see Appendix \ref{sec:diff_class_details}.
Its performance is very good ($84\%$ accuracy) on in-distribution data (i.e.~the validation split), but it worsens on a different test set composed of official JLPT past exam sentences ($38\%$ accuracy). Our hypothesis is that the latter test set contains very long sentences composed of many relative clauses, which are very different from the sentences used for training.
% However, during internal testing it was found to work well enough. That was partially confirmed by the raters evaluation of difficulty for the retrieval system. Nevertheless, we explore possible improvements in the conclusion (Section \ref{sec:conclusion}).

% websites eg https://nihongokyoshi-net.com/ https://jlptsensei.com/ and others.

\subsubsection{Sense embeddings}
%\paragraph{Sense embeddings}
%  Then, the embeddings of the target word in the context sentence and in each candidate sentence are computed using the sense embedding model. From that, we obtain a similarity score from comparing cosine similarity between the target word in the context sentence and the candidate sentence.
We use MirrorWiC~\cite{liu_mirrorwic_2021} to fine-tune multiple baseline PLMs with 10,000 sentences randomly chosen from our corpus.
To guide model selection, we look at their performance on two WiC tasks, XL-WiC~\cite{raganato-etal-2020-xl} and AM2iCo~\cite{liu-etal-2021-am2ico}. MirrorWiC fine-tuning shows a small improvement on both tasks for BERT-base-japanese, over the same base model and a Japanese Sentence Transformer.\footnote{\href{https://huggingface.co/sonoisa/sentence-bert-base-ja-mean-tokens-v2}{sonoisa/sentence-bert-base-ja}}

% added that it is the standard
To obtain the embeddings, we average the last $4$ layers of the embedding model, and across the sub-tokens that make up the target word, following \citet{liu_mirrorwic_2021}.

\section{Evaluation}
%In this section,  we describe our chosen evaluation strategy on sentence suggestion systems designed for language learning, along with its objectives and setup.

\subsection{Goals of the evaluation}
% the research questions
We outline the core research questions that guide our investigation.

% old version
% (Q1) How beneficial are sentence suggestion systems for language learners today?
% (Q2) How do the automated quality metrics we used to guide the development of the retrieval system compare with human judgment?
% (Q3) How good are PLMs at instruction following for such a complex task with many requirements?
% (Q4) Is text retrieved from a corpus (assumed to be human-authored) preferred to generated text?
% (Q5) Can a PLM judge open ended text generation quality in the language learning field reliably, and how does it compare with human ratings?
\begin{enumerate}[start=1,label={\bfseries Q\arabic*:}, nosep]
\item The capabilities of LLMs such as GPT-4 in rating text have been explored \cite{chen-etal-2023-exploring-use}. Therefore, can GPT-4 evaluate the quality of Japanese sentences from the perspective of L2 learners, and how do its assessments compare to those given by humans?

\item How do the automated quality metrics we used to guide the development of the retrieval approach compare with human judgment?

\item How good are PLMs at following instructions for this complex task?

\item Is text retrieved from a corpus (assumed to be human-authored) preferred to generated text?

\item What do humans think of their output? % What are their comments? %Their comments could provide insights on what the shortcomings and possible solutions are.
\end{enumerate}

We try to answer those questions by asking volunteer L2 learners and Japanese native speakers to manually rate and rank systems outputs.

\subsection{Selected baselines}
% the retrieval, and the llms (retrieval discussed in previous section, as well as the llms
The systems we consider are the retrieval approach (Section~\ref{sec:retr}), LLM-jp, a Japanese PLM,\footnote{\href{https://huggingface.co/llm-jp/llm-jp-13b-instruct-full-jaster-dolly-oasst-v1.0}{llm-jp/llm-jp-13b-instruct-full-jaster-dolly-oasst-v1.0}} and GPT-3.5.
Specifically, throughout the paper, when mentioning GPT-3.5 we mean GPT-3.5-turbo-0613, while GPT-4 is GPT-4-0125-preview.

\subsection{Evaluation data preparation}
We build a set of target words from those used in the human evaluation of \citet{tolmachev-kurohashi-2017-automatic} and also add words from a work in WSD by \citet{okumura_semeval-2010_2011}.
The former study involved 14 target words, and the latter 50, sharing one word, resulting in a total of 63. We randomly divided them into 53 for validation and experimental use, and 10 for testing and human evaluation, but ensuring a test set composition of 3 nouns, 4 verbs, 2 adjectives, and 1 adverb.

In addition, for every target word, we obtain a context sentence by randomly selecting sentences from \textit{yourei} and \textit{gogo},\footnote{\href{https://yourei.jp}{https://yourei.jp}, \href{https://dictionary.goo.ne.jp}{https://dictionary.goo.ne.jp}} websites which provide a search engine for snippets of text content. 

\subsection{Human evaluation guidelines}
% We take the words for human evaluation described earlier $w_1, w_i, ..., w_N$, along with their associated context sentence $s_{0i}$.

We consider as a query the input for the task (Equation \ref{eq:task}), namely the selected word for human evaluation, along with their associated context sentence and target level.
In this experiment we target levels N1, N3 and N5. 
The system outputs are randomly ordered and presented with the query, forming an ``annotation block''. Each baseline provides $K=5$ sentences.
This results in 30 blocks ($10$ queries $\times \  3$ levels), and $150$ sentences for each system ($30$ blocks $\times \  5$ output sentences). 

We ask evaluators to rate:
\begin{enumerate}[nosep]
\item \textbf{Difficulty level}, by rating the difficulty of each sentence on the JLPT scale. %, which ranges from N1 (advanced level) to N5 (beginner level). 
 This is to see how closely systems match the target difficulty.
\item 
\textbf{Sense similarity}, by evaluating whether the usage of the target word in each sentence aligns with its sense in the original context. %, using broad and intuitive judgment rather than strict dictionary definitions.
% eg aite has different things
% this is very hard and could be unclear among people
 This is to see whether the proposed sentences retain the use of the word in a similar sense.%, and to see whether different raters tend to give different responses.

\item \textbf{Rejection}: sentences should be marked for rejection if they are deemed not useful (e.g. unnatural usage) or confusing (e.g., grammatical or segmentation errors). %, or for any other specified reasons. A comment should accompany the rejection to clarify the rationale.
\item \textbf{Syntactic Diversity}, by examining the variety in sentence structure and the different grammatical constructions used to incorporate the target word. %This evaluation focuses on identifying the range of syntactic applications to ensure a broad learning experience.
%This is to evaluate how diverse the groups of sentences are.

\item \textbf{System Ranking}: after rating each system's outputs, rank them from best to worst. The ranking should consider the overall utility for language learners at the target proficiency.
\end{enumerate}
% We perform an initial test for agreement where we explained the rating guidelines and deemed it okay to pursue the complete evaluation.
We demonstrated the task and explained the evaluation guidelines. %Annotating all items required about 2 hours and 30 minutes, but some evaluators only rated a subset of the annotation blocks. -- maybe not too relevant
The total participants are 5, of which 3 are native Japanese speakers and 2 are learners of proficiency N1-N2.
%All are also proficient in English.
Addionally, we include an annotation block example in Appendix~\ref{sec:evaluation form}.
% human native  1    Aizawa
% human native  2    Takizawa
% human native  3    Tsurusaki
% human learner 1    Henri
% human learner 2    Willy
% names should stay anonymous

\subsection{PLM evaluation protocol}
We feed GPT-4 a modified version of the evaluation guidelines, the system outputs, and ask it to rate them. More details can be found in Appendix \ref{sec:gpt4_eval_prompt}.
Empirically, we noticed that ratings for the same prompt sometimes were different, even when trying to reduce variability.
So, we query GPT-4 three times, and also obtain its majority vote. We note that in some cases this could still result in an unclear rating. % in that case we select as best system the leftmost one (it is random)

\section{Results and Discussion}
% Present the results and compare
% which system got the most votes and ranks
% how about sense and the shortcomings
% how about the difficulty and matching of it by retrieval / llms etc (should be a table + a discussion in general)
%In this section, we present the results from the human evaluation and the evaluation carried with GPT-4.
In this section, we present the results from the evaluation.
We try to address our research questions in three main parts: agreement between raters; systems comparison; comments and error analysis.

\subsection{Q1: Agreement of ratings}
%\paragraph{Q1: Agreement of ratings}
The Intraclass correlation coefficient (ICC) is a widely used statistical measure for reliability, that reflects the degree of correlation and agreement between ratings \cite{koo_guideline_2016}. 
The reason for choosing this metric is that it takes into account the magnitude of the differences between scores. 
For example, it is important that if a sentence is rated N1 by one person and N5 by another, it is seen as a larger disagreement than one rated N1 and N2.

%Calculating the agreement on one rated quality at a time does not take into account the fact that while rating sentences, evaluators could be influenced by the other previously given ratings.

We compute the metric with the pingouin library,\footnote{\href{https://pingouin-stats.org/build/html/index.html}{https://pingouin-stats.org/build/html/index.html}} and we convert ratings from ordinal labels into numbers, mapping them in a scale where the relative distances are the same among labels. %(eg N1 N2 N3 N4 N5 -> 1 0.75 0.5 0.25 0) - using other scales does not change the results much
Following \citet{hackl_is_2023}, who studied the reliability of GPT-4 in a similar experiment, we use a specific ICC setting based on a two-way mixed effect model. 
In short, ICC(3,1), according to the naming convention of \citet{shrout_icc}.
% because we suppose that each rater is different, and we do not plan on taking the average of ratings. using the average ratings makes it go higher.

\subsubsection{GPT-4 rating consistency}
%\paragraph{GPT-4 rating consistency}
In Table \ref{tab:icc}, we report ICC values for the quality ratings across groups of raters. % also the pairwise for level / ranking ? need to see the others
We include in this table only raters who compiled at least half of the blocks for each target level, in order to have a generalizable idea of the agreement. % for all difficulty levels.

\begin{table*}[ht]
\centering
\small
\smallskip\noindent
\resizebox{0.85\linewidth}{!}{%
\begin{tabular}{l|rr|rr|rr}
\toprule
\textbf{Rater group}$\rightarrow$ & \multicolumn{2}{c}{GPT-4 ($N=3$)} & \multicolumn{2}{|c|}{Human ($N=3)$} & \multicolumn{2}{|c}{All ($N=4$)} \\
\midrule
\textbf{Rated item}$\downarrow$ & \textbf{ICC(3,1)} & \textbf{95\% CI} &\textbf{ICC(3,1)} & \textbf{95\% CI} & \textbf{ICC(3,1)} & \textbf{95\% CI} \\
\midrule
Level & 0.941 & [0.93, 0.95] & 0.681 & [0.63, 0.73] & 0.673 & [0.63, 0.72] \\
Sense & 0.640 & [0.59, 0.68]  & 0.258 & [0.18, 0.33] & 0.108 & [0.06, 0.17]\\
Reject & 0.861 & [0.84, 0.88]  & 0.238 & [0.18, 0.30] & 0.244 & [0.20, 0.30]\\
Syn. diversity & 0.778 & [0.70, 0.84] & 0.214 & [0.08, 0.36] & 0.236 & [0.13, 0.36]\\
Ranking & 0.694 & [0.60, 0.78] & 0.218 & [0.09, 0.36] & 0.218 & [0.12, 0.34] \\ 
\bottomrule
\end{tabular}
}
\caption{ICC estimates and their $95\%$ confidence intervals (CI) for different groups. $N$ indicates the number of raters in the group. In the last group, we consider the humans and the majority vote of GPT-4.}
\label{tab:icc}
\end{table*}

\begin{table*}[ht!]
\centering
\small
\smallskip\noindent
\resizebox{0.85\linewidth}{!}{%
    \begin{tabular}{l|lllllllll}
    \toprule
    \textbf{Rater}$\downarrow\rightarrow$ & GPT-4$_{\text{majority}}$ & GPT-4$_1$ & GPT-4$_2$ & GPT-4$_3$ & HL 1 & HL 2 & HN 1 & HN 2 & HN 3 \\
    \midrule
    GPT-4$_{\text{majority}}$ & 1 & 0.80$^*$ & 0.78$^*$ & 0.93$^*$ & 0.37$^*$ & 0.22$^*$ & 0.37$^*$ & 0.05 & 0.20 \\
    GPT-4$_1$  & 0.80$^*$ & 1 & 0.55$^*$ & 0.72$^*$ & 0.33$^*$ & 0.17 & 0.35$^*$ & 0.02 & 0.11 \\
    GPT-4$_2$ & 0.78$^*$ & 0.55$^*$ & 1 & 0.82$^*$ & 0.29$^*$ & 0.17 & 0.45$^*$ & 0.13 & 0.28$^*$ \\
    GPT-4$_3$ & 0.93$^*$ & 0.72$^*$ & 0.82$^*$ & 1 & 0.37$^*$ & 0.21$^*$ & 0.28$^*$ & -0.03 & 0.20 \\
    HL 1 & 0.37$^*$ & 0.33$^*$ & 0.29$^*$ & 0.37$^*$ & 1 & 0.29$^*$ & 0.46$^*$ & 0.13 & 0.68$^*$ \\
    HL 2 & 0.22$^*$ & 0.17 & 0.17 & 0.21$^*$ & 0.29$^*$ & 1 & 0.22$^*$ & 0.14 & 0.47$^*$ \\
    HN 1 & 0.37$^*$ & 0.35$^*$ & 0.45$^*$ & 0.28$^*$ & 0.46$^*$ & 0.22$^*$ & 1 & 0.30$^*$ & 0.42$^*$ \\
    HN 2 & 0.05 & 0.02 & 0.13 & -0.03 & 0.13 & 0.14 & 0.30$^*$ & 1 & 0.42$^*$ \\
    HN 3 & 0.20 & 0.11 & 0.28$^*$ & 0.20 & 0.68$^*$ & 0.47$^*$ & 0.42$^*$ & 0.42$^*$ & 1 \\
    \bottomrule
    \end{tabular}}
    \caption{Pairwise agreement matrix of ICC(3,1) scores on \textbf{ranking preferences}. ``HL'' refers to a human learner, while ``HN'' to a human native speaker. $^*$: \textit{P}-value is less than $.05$.}
    \label{tab:rank_pair_agreement}
\end{table*}

% TABLE WITH RANKINGS
\begin{table*}[h!]
\centering
\small
\smallskip\noindent
\resizebox{0.85\linewidth}{!}{%
\begin{tabular}{l|rrrr|rrrr|rrrr}
\toprule
\textbf{System}$\rightarrow$ & \multicolumn{4}{|c|}{Retrieval} & \multicolumn{4}{|c|}{LLM-jp} & \multicolumn{4}{|c}{GPT-3.5} \\
\midrule
\textbf{Rater}$\downarrow$, \textbf{Target}$
\rightarrow$ & N1 & N3 & N5 & Tot. & N1 & N3 & N5 & Tot. & N1 & N3 & N5 & Tot. \\
\midrule
GPT-4$_{\text{majority}}$ & \textbf{7} & \textbf{5}& \textbf{5} & \textbf{17} & 2 &	2 & 2 & 6 & 1 & 3 & 3 & 7 \\
HL 1$^\dagger$ & \textbf{5} & \textbf{4} & -- & \textbf{9} & 0 & 0 & -- & 0 & 0 & 2 & --	& 2 \\
HL 2 & \textbf{4} & 3 & \textbf{6} & \textbf{13} & 2 & 2 & 2 & 6 & \textbf{4} & \textbf{4} & 2 & 10 \\
HN 1 & \textbf{10} & \textbf{4} & \textbf{10} & \textbf{24} & 0 & 2 & 0 & 2 & 0 & \textbf{4} & 0 & 4 \\
HN 2 & \textbf{7} & 1 & \textbf{8} & \textbf{16} & 2 & \textbf{5} & 1 & 8 & 1 & 4 & 1	& 6\\
HN 3$^\dagger$ & \textbf{7} & 1 & -- & \textbf{8} & 1 & 2 & -- & 3 & 0 & \textbf{6} & -- & 6\\
\bottomrule
\end{tabular}}
\caption{Number of annotation blocks in which the considered baseline is rated first in overall quality, by target difficulty level. $^\dagger$: The participant mostly rated blocks with target level N1 and N3 only, because of time constraints.}
\label{tab:rankings-per-level}
\end{table*}

\begin{figure*}[h!]
    \centering
    \smallskip\small
    \includegraphics
    %[max size={\textwidth}{0.42\textheight}]
    [width=0.85\linewidth]
    {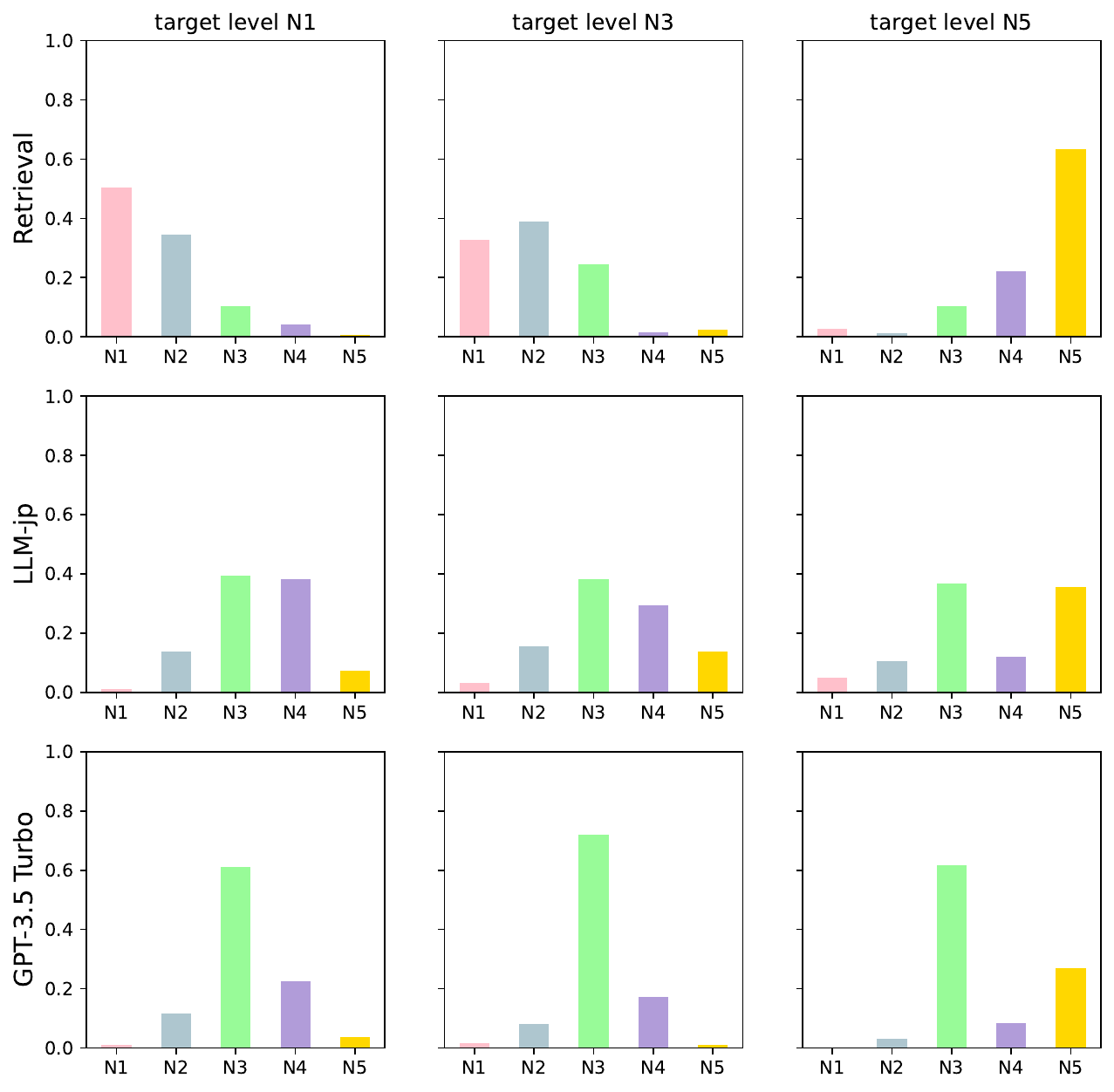}
    \caption{Evaluators' ratings on difficulty. Each row presents the proportions of JLPT labels assigned by humans for one system, across the three target difficulty levels set for the evaluation.}
    \label{fig:label_diff}
\end{figure*}

For GPT-4, despite setting its behavior to be nearly deterministic and obtaining ratings on the same day, we observed that the consistency of its ratings varies by type. The model shows excellent agreement in assessing JLPT levels and good consistency in rejecting sentences. However, its consistency is lower for other evaluation areas like sense similarity, syntax diversity, and model ranking. Using a mean combination of ratings improves consistency, but comes at the cost of more forward passes on the same long inputs.
A way to further mitigate this is improving the prompt.
% not shown in table, ICC(3,k) - mean ratings
% as a general guideline: 
%\cite{koo_guideline_2016}
% Values less than 0.5 are indicative of poor reliability, values between 0.5 and 0.75 indicate moderate reliability, values between 0.75 and 0.9 indicate good reliability, and values greater than 0.90 indicate excellent reliability.
% but the interpretation should be data-domain-task specific.

\subsubsection{Agreement among groups}
%\paragraph{Agreement among groups}
Focusing on human raters, it seems that agreement on qualities except difficulty level is quite low (Table \ref{tab:icc}).
One reason for this could be that the guidelines for other metrics are too generic, which causes more variability in the ratings. However, we expected that language learners and native speakers may not have the same rating patterns.
Additionally, since we required many ratings at once, there could be some additional effects at play, such as fatigue or bias from the order of annotation. %it could also be hard to make the same decision due to the complex nature 
%therefore, we could devise a better experimental condition next time..

\subsubsection{Pairwise agreement on ranking}
%\paragraph{Pairwise agreement on ranking}

To further investigate whether GPT-4 ranks similarly to humans,
 in Table \ref{tab:rank_pair_agreement} we report the pairwise agreement for the preferred system ranking from all annotators. 
 
Inter-rater agreement between GPT-4 and humans is generally lower than those among humans of different groups. This suggests that humans, regardless of whether they are native speakers or not, have more similar ranking preferences compared to the AI models. 
However, there are also outliers, such as HN2, who has a way of ranking that shows no agreement with many other raters.
This highlights the challenge in aligning AI evaluations with human preferences and confirms that, even among humans, there is significant disagreement on judging learning material suitability.

%\subsection{Q2-Q3-Q4: Quantitative analysis of ratings}
%shortening
%\subsection{Q2\char`\~Q4: Quantitative analysis of ratings}
\subsection{Q2-3-4: Quantitative analysis of ratings}

% Q2 How do the automated quality metrics we used to guide the development of the retrieval approach compare with external judgment?
% Q3 How good are PLMs at instruction following for such a complex task with many requirements?

%[together, we analyze the ratings given]
%[level -> classifier quite good]
After the agreement analysis, we discuss how raters evaluated the systems. %We report the combined scores assigned by all the human participants for the three systems.
For qualities other than difficulty and ranking preference, we report the main empirical findings in the following, and release additional figures in Appendix \ref{sec:additional plots}.
% the plots are very similar, so it would be too long to analyze in detail every group. Also, human opinions are the main reason we did the human evaluation, and also previously we saw that gpt is kind of similar in giving ratings (at least for levels). The point is that there is inherent disagreement among humans.
\subsubsection{Difficulty level ratings}
%\paragraph{Difficulty level ratings}

Figure \ref{fig:label_diff} shows the proportion of human-assigned JLPT difficulty labels for each system, grouped by target level.
When considering how close the difficulty of proposed sentences is to the target level, our retrieval approach is markedly better for N1 and N5, while for N3, it produced a significant proportion of harder sentences.
GPT-3.5 seems better for N3, but being so consistent is not always an advantage because it makes it difficult to adapt to user requirements, for example when requesting advanced sentences. LLM-jp also had issues following the prompt: repetitions, sentences without the target word, incoherent text.

\subsubsection{Sense similarity ratings}
%\paragraph{Sense similarity ratings}
When the raters indicated whether the target word in each sentence had a similar meaning as the one in the context, the vast majority classified the sense as being the same. The percentage of sentences rated as ``not similar'' was only about $2\%$ for the retrieval, and $13\%$ for the generative baselines. This shows that the systems generally succeed in producing examples with similar nuances.

\subsubsection{Rejection ratings}

According to our evaluation guidelines, unnatural sentences and those with confusing errors should be marked.
On average, $8\%$ of sentences suggested by the retrieval were rejected, while for LLM-jp it was $13\%$, and $16\%$ for GPT-3.5.

Checking raters' comments confirmed that there were some segmentation errors in retrieval and generation baselines, such as sentences starting with punctuation, or with a fragment.
It seems that generative models are more prone to errors, while the retrieved sentences are better in this aspect ``by design''.
Still, careful text pre-processing and post-processing is needed as sentences with errors can be confusing for beginner learners.

For a couple of concrete examples, the following sentence from the retrieval approach was rejected by some human participants because it sounded unnatural and too literary:
\begin{CJK}{UTF8}{min}
    この闘いは今日の場合では大概は容易ならぬ苦闘だからだ。
    \end{CJK} \textit{``As for this fight, in today's situation, it is generally a difficult struggle''}.

Finally, the following was generated by LLM-jp and was rejected because of the presence of confusing characters and English words at the beginning:
    \begin{CJK}{UTF8}{min}
    \texttt{favorite dish is sushi.1.}右手で持っていたスプーンを左手でも持てるようになったんだ。
    \end{CJK} \textit{``The spoon that I was holding with the right hand, I became able to hold with the left hand as well''}.

\subsubsection{Diversity ratings}
%\paragraph{Diversity ratings}
Considering the syntax diversity of the list of sentences, the retrieval method earned the most ``high'' ratings across all target levels. GPT-3.5 received mostly ``medium'' votes, and LLM-jp got the lowest. The latter model often produced repetitive sentences, where only one or two words would differ between each generated sentence. This highlights another issue in zero-shot generation, i.e.~that it is difficult to have both diversity and adherence to instructions.

\subsubsection{System ranking ratings}

%\paragraph{System ranking ratings}
%Q4 Is text retrieved from a corpus (assumed to be human-authored) preferred to generated text? AKA
% [ranking analysis] - Q4

Table \ref{tab:rankings-per-level} presents votes on system ranking by human participants and GPT-4.
The sentence lists produced by the retrieval system are the best overall for all raters when considering the total vote count. Except for HL2 and HN3, the retrieval system is rated best in over $50\%$ of cases.
When considering target levels, it also markedly wins in suggesting lists for advanced and beginner target difficulty levels, while it is not rated best as much for the intermediate level.
The sentences suggested by the retrieval system for N3 are often on the more difficult end, as shown in Figure \ref{fig:label_diff}.
% remove this table (redundant now
% \begin{table*}[h]
% \centering
% \begin{tabular}{l| rrr |r}
% \toprule
% \textbf{Rater}$\downarrow$, \textbf{System}$\rightarrow$ & Retrieval & LLM-jp & ChatGPT & \textbf{Total} \\
% %rank_human & 1st & 1st & 1st &  \\
% \midrule
% %gpt4 & 17 & 6 & 7 & 30 \\
% GPT-4$_1$ & 18 & 4 & 8 & 30 \\
% GPT-4$_2$ & 17 & 5 & 8 & 30 \\
% GPT-4$_3$ & 16 & 6 & 8 & 30 \\
% Human learner  1$^\dagger$ & 9 & 1 & 2 & 12 \\
% Human learner  2 & 13 & 6 & 10 & 29 \\
% Human native   1 & 24 & 2 & 4 & 30 \\
% Human native   2 & 16 & 8 & 6 & 30 \\
% Human native  3$^\dagger$ & 8 & 3 & 6 & 17 \\
% \bottomrule
% \end{tabular}
% \caption{Number of instances where a system was rated best by participants in the study, by target difficulty level. The ``Total'' column refers to the total amount of blocks completed by participants. $^\dagger$: Mostly rated blocks with target level N1 and N3, because of time constraints.}
% \label{tab:best_votes}
% \end{table*}

\subsection{Q5: Qualitative analysis and participants comments}

% moved to previous section on reject analysis

% Careful pre-processing and post-processing of text is needed before actually presenting the sentences, especially to beginner learners, as these errors could be confusing.

A native speaker commented on a target word in the evaluation (\begin{CJK}{UTF8}{min}全然\end{CJK}, \textit{zenzen}). It is commonly used in negative statements, to mean ``not at all'' \cite{Sawada2007TwoTO}.
Using it in positive statements can be considered ``slightly broken'' in formal situations, but it was correct a hundred years ago, and it is used in today's slang. 
In that case, GPT-3.5 produced a similar sentence as the context in which the usage was ``uncommon''. 
Indeed, the context sentence was from an excerpt of a work published in 1938 by Osamu Dazai, a famous Japanese writer.
This should prompt thinking about what actually makes a correct sentence. %?
% This usage ties into a linguistic principle known as polarity, a concept found across all human languages \cite{lobner_polarity_2000}.
% When a word typically associated with negative contexts is used in a positive statement, it can sound odd, akin to saying ``I ran at all" in English.
%too long otherwise
Language learners noted that many sentences contained one or two difficult kanji, encountered at higher proficiency levels, even though the overall sentence structure is more straightforward to understand.
This happened mostly with the retrieval approach, which did not take word difficulty explicitly into account.

\section{Conclusion}

This paper outlines a methodology for suggesting example sentences to learners of Japanese. It is adaptable to other languages with minor adjustments.
The baselines we consider highlight many possible roles of PLMs: assessing difficulty, encoding semantic representations, directly producing sentences and evaluating their quality, all of which could be investigated further on the basis on their applicability in AI-supported language learning and other fields in education technology.

From the feedback and data collected from the human evaluation, we can point out the potential for improving and combining these systems to balance their shortcomings, even though the retrieval methodology was considered to be the best in terms of diversity and adherence to difficulty level.

The challenge of evaluating generated text prompted us to explore a state-of-the-art LLM's ability in rating sentence quality. In our opinion, it is a promising direction because the model seems to be able to evaluate linguistic features of sentences. We found good agreement in rating text difficulty, but since each person could make different assessments, finding a way to take that variability into account could be useful for personalization.

It could be studied whether using word-level features can prevent unknown kanji from appearing in example sentences. Such features could be JLPT labels or the school grade level they are taught in. 
%[Jōyō footnote - https://www.kanshudo.com/collections/joyo_kanji]
Another research challenge is estimating the real vocabulary known by the learner, modeling the process of second language acquisition \cite{settles-etal-2018-second, cui-modeling}. Additionally, there is potential for suggestion and generation of material based on each learner's interests.

A direction to explore further is to experiment with more advanced LLM prompting strategies, such as chain of thought or reinforcement learning, to iteratively refine outputs for better adaptation to learners' preferences. A retrieval approach like ours could serve as a starting point.

%We hope our findings and collected feedback will help researchers in making systems for obtaining L2 learning material in a way that more closely benefits language learners.

\section*{Limitations}

In our work, the retrieval approach scores sentences using mainly unsupervised approaches and PLMs.

The corpus we build is not as large as other corpora. 
In our comparisons, for LLMs we explored only basic prompting strategies without fine-tuning, wanting to investigate approaches in a setting without labeled data.

As for the evaluation, the number of volunteers who participated in the study was quite limited and the agreement values are not very high, indicating that the results are not generalizable to larger groups. Nevertheless, we believe that the feedback and guidelines could be valuable for future research.
About half of them were foreign students, and their feedback was valuable. Unfortunately, due to lack of resources, none of the native speakers were language educators. Involving language teachers would be advisable.
Additionally, comparing our baselines with the approach of \citet{tolmachev-kurohashi-2017-automatic} would have been insightful. However, due to the absence of a practical implementation and limited resources for human evaluation, we opted for PLM baselines.

\section*{Ethics Statement}
Because of the training methods of base LLMs, sentences generated or retrieved using these approaches could reflect negative biases that could impact or influence negatively the model of language that is internalized by the learners.
It poses an increased risk when there are not enough sources of information, or limited sharing of ideas and communication with other learners and native speakers of the foreign language that can more effectively teach distinguishing polite and casual register and other aspects of pragmatics, other than just word usage.

% Scientific work published at ACL 2023 must comply with the ACL Ethics Policy.\footnote{\url{https://www.aclweb.org/portal/content/acl-code-ethics}} We encourage all authors to include an explicit ethics statement on the broader impact of the work, or other ethical considerations after the conclusion but before the references. The ethics statement will not count toward the page limit (8 pages for long, 4 pages for short papers).

\section*{Acknowledgements}
This work was supported by JSPS KAKENHI Grant Number 24K03231.
We thank the LLM Research and Development Centre for LLMs, National Institute of Informatics, Japan, for the support.

We also wish to thank the volunteer participants in the evaluation experiments for their help, Arseny Tolmachev for the precious advice, and the anonymous reviewers for their feedback.

% Entries for the entire Anthology, followed by custom entries
\bibliography{anthology,custom}
\bibliographystyle{acl_natbib}

\onecolumn
\appendix

\section{Difficulty classifier training and evaluation}
\label{sec:diff_class_details}

% insert the training details, the confusion matrix and classification report ? 
\begin{table}[ht!]
\centering
\begin{tabular}{l|l}
\toprule
\textbf{Parameter}                  & \textbf{Value}           \\
\midrule
model                               & cl-tohoku/bert-base-japanese-v3   \\ \hline
tokenizer                           & model's AutoTokenizer    \\ \hline
no. labels                         & 5 \textit{(N1, N2, N3, N4, N5)}                       \\\hline

learning rate                      & 2e-5                      \\ \hline

batch size     & 8                         \\ \hline
no. epochs                  & 10                        \\ \hline
adam $\beta_1$                       & 0.9                       \\ \hline
adam $\beta_2$                       & 0.999                     \\ \hline
adam $\epsilon$                        & 1e-7                      \\ \hline
weight decay                       & 0.01                      \\ 
\bottomrule
\end{tabular}
\caption{Summary of training parameters for the difficulty classifier.}
\label{table:training_params}
\end{table}

% class report for the same distribution data
\begin{table}[ht!]
\centering
\begin{tabular}{c|ccc|c}
\toprule
\textbf{Class} & \textbf{Precision} & \textbf{Recall} & \textbf{F1-score} & \textbf{Support} \\ \hline
N5             & 0.88               & 0.88            & 0.88              & 25               \\ \hline
N4             & 0.90               & 0.89            & 0.90              & 53               \\ \hline
N3             & 0.78               & 0.90            & 0.84              & 62               \\ \hline
N2             & 0.71               & 0.79            & 0.75              & 47               \\ \hline
N1             & 0.95               & 0.77            & 0.85              & 73               \\ \midrule
\multicolumn{1}{r|}{\textbf{Macro Avg}} & 0.84               & 0.84            & 0.84              & 260              \\ \hline
\multicolumn{1}{r|}{\textbf{Weighted Avg}} & 0.85               & 0.84            & 0.84              & 260              \\ \hline
\multicolumn{1}{r|}{\textbf{Accuracy}} & 0.84 & &  & 260              \\  \bottomrule
\end{tabular}
\caption{Metrics on data from the test split from the same data distribution for the difficulty classifier.}
\label{table:classification_report_same}
\end{table}

% class report for the test data from JLPT official exams
\begin{table}[ht!]
\centering
\begin{tabular}{c|ccc|c}
\toprule
\textbf{Class} & \textbf{Precision} & \textbf{Recall} & \textbf{F1-score} & \textbf{Support} \\ \hline
N5             & 0.62               & 0.66            & 0.64              & 145               \\ \hline
N4             & 0.34               & 0.36            & 0.35              & 143               \\ \hline
N3             & 0.33               & 0.67            & 0.45              & 197               \\ \hline
N2             & 0.26               & 0.20            & 0.23              & 192               \\ \hline
N1             & 0.59               & 0.08            & 0.15              & 202               \\ \midrule
\multicolumn{1}{r|}{\textbf{Macro Avg}} & 0.43               & 0.39            & 0.36              & 879              \\ \hline
\multicolumn{1}{r|}{\textbf{Weighted Avg}} & 0.43              & 0.39            & 0.36              & 879              \\ \hline
\multicolumn{1}{r|}{\textbf{Accuracy}} &  0.38 & &  & 879              \\  \bottomrule
\end{tabular}
\caption{Metrics on a test set of sentences from the official JLPT exams for the difficulty classifier.}
\label{table:classification_report_out}
\end{table}

\begin{figure}
    \centering
    \includegraphics[max size={\textwidth}{0.38\textheight}]{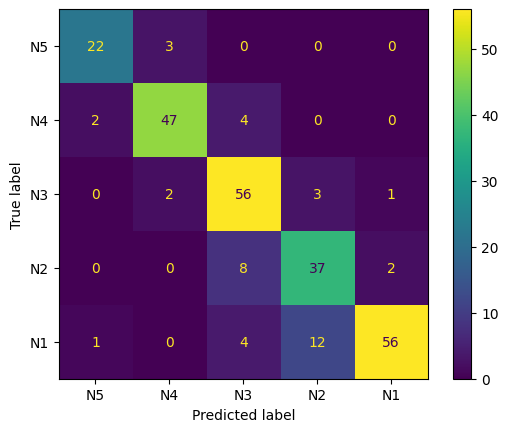}
    \caption[Confusion matrix for the difficulty classifier on same-distribution data]{Confusion matrix for the difficulty classifier, on sentences obtained in the same way as the training data (i.e. distant supervision labeling from language websites).}
    \label{fig:conf_test_same_dist}
\end{figure}

\begin{figure}
    \centering
    \includegraphics[max size={\textwidth}{0.38\textheight}]{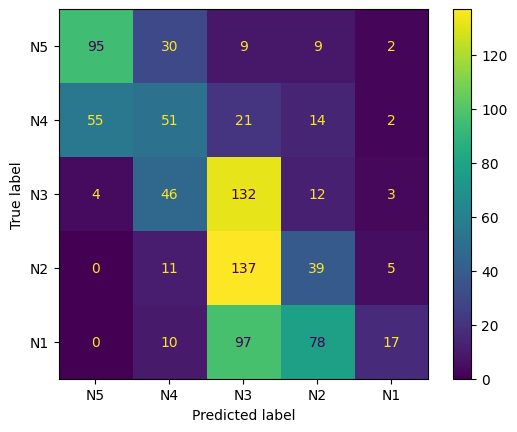}
    \caption[Confusion matrix for the difficulty classifier on out-of-distribution data]{Confusion matrix for the difficulty classifier, on sentences obtained from a different source (i.e. past exams from the official JLPT website).}
    \label{fig:conf_test_official}
\end{figure}

% \section{Sense embedding model selection}
% \label{sec:mirrorwic_details}
\newpage

\section{Human evaluation form - Example of evaluation block}
\label{sec:evaluation form}

\begin{figure}[h!]
    \centering
    \includegraphics[angle=90,origin=c, max size={\textwidth}{0.68\textheight}]{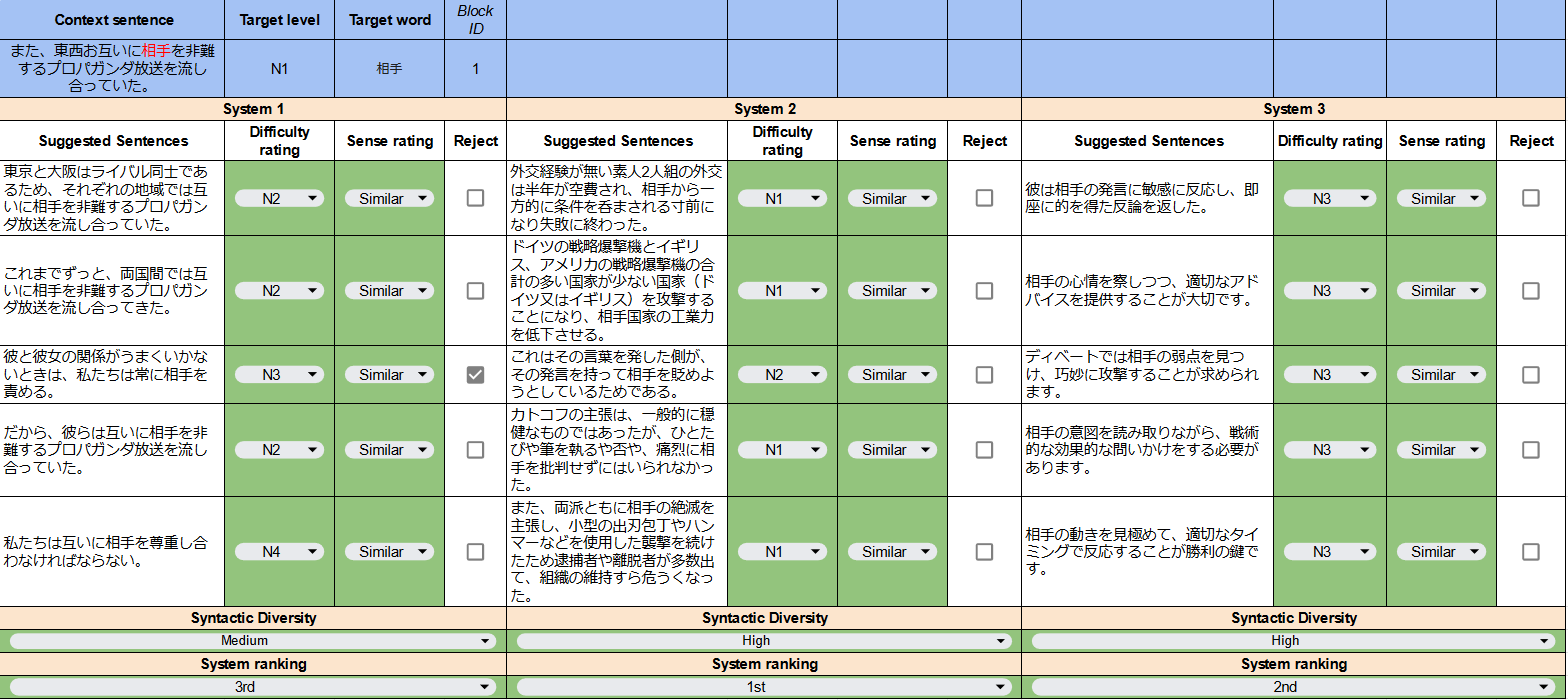}
    \label{fig:block_example}
\end{figure}

\newpage

\twocolumn
\section{LLM baselines prompts}
\label{sec:prompts}

We share the prompts, obtained with manual testing and trial and error. We found that the models responded in a satisfactory way also to prompts where the request was formulated in plain English, as well as in Japanese.

For LLM-jp, this was the prompt used to obtain the final outputs:
\begin{quote}
\begin{tt}

    write \textit{k} \textit{target level} example sentences in japanese, that must contain the word "\textit{target word}" used in a similar sense as "\textit{context sentence}". following are \textit{k} diverse sentences that must use "\textit{target word}": 
\end{tt}
\end{quote}
For GPT-3.5, we used the same prompt as the other LLM, and only appended the following instruction to reduce verbosity.
\begin{quote}
\begin{tt}
    Provide sentences in Japanese in a numbered list, without any translation or romaji.
\end{tt}
\end{quote}

\section{GPT-4 evaluation prompt}
\label{sec:gpt4_eval_prompt}
We present the prompt given to GPT-4 when rating evaluation blocks with the baselines outputs:

\begin{quote}
\begin{tt}
    This evaluation aims to rate and compare three systems in providing good example sentences for learners of Japanese at different proficiency levels.
An annotation block consists of proposed sentences by 3 systems for a target word, a context sentence and a target difficulty level.
The lists of sentences are supposed to help language learners to see diverse examples of a target word in context.

Difficulty: Rate the difficulty of each sentence according to the JLPT (Japanese Language Proficiency Test) scale, where N1 is the most difficult and N5 is the easiest.
Indicate which level a sentence belongs to (one of N1, N2, N3, N4, N5). It is possible that for the target level, the system proposes a sentence that is of a different level (higher or lower).
Below is a summary of the proficiency levels.\footnote{Taken from \href{https://www.jlpt.jp/e/about/levelsummary.html}{https://www.jlpt.jp/e/about/levelsummary.html}. The description are put into a table for readability.}

\begin{center}
\centering
\small
\smallskip\noindent
    \begin{tabular}{c|p{0.3\textwidth}}
\toprule
\textbf{Level} & \textbf{Description}                                                                                         \\ \midrule
N1             & Complex and abstract Japanese across various contexts.                                                               \\ \hline
N2             & Everyday Japanese in varied situations, with clear materials on different topics.                                     \\ \hline
N3             & Japanese in common everyday situations.                                                                               \\ \hline
N4             & Basic Japanese understanding, including familiar topics, basic vocabulary, and kanji.                                 \\ \hline
N5             & Fundamental Japanese, including hiragana, katakana, and basic kanji.                                                  \\ \bottomrule
\end{tabular}
\end{center}

Sense Similarity: Indicate if the target word in each sentence maintains a close sense as in the original context. Possible values: "similar", "not similar".
Think broadly and intuitively, rather than strictly by dictionary definitions.

Reject: For each sentence, indicate "Reject" if you think the sentence is not good or useful (for example because it does not reflect natural use).

Sentence diversity: For each system output list, rate the sentences diversity, focusing on the amount of different uses of syntax and structure. Possible values: "Low", "Medium", "High".

System ranking: Rank the systems' outputs from best to worst, considering the overall usefulness of the example sentences for that word, for a language learner of that proficiency level.

Comment: Leave a short comment.
\end{tt}
\end{quote}

\newpage
\onecolumn

\section{Additional rating statistics}
\label{sec:additional plots}

\begin{figure}[ht!]
    \centering
    \includegraphics[max size={\textwidth}{0.42\textheight}]{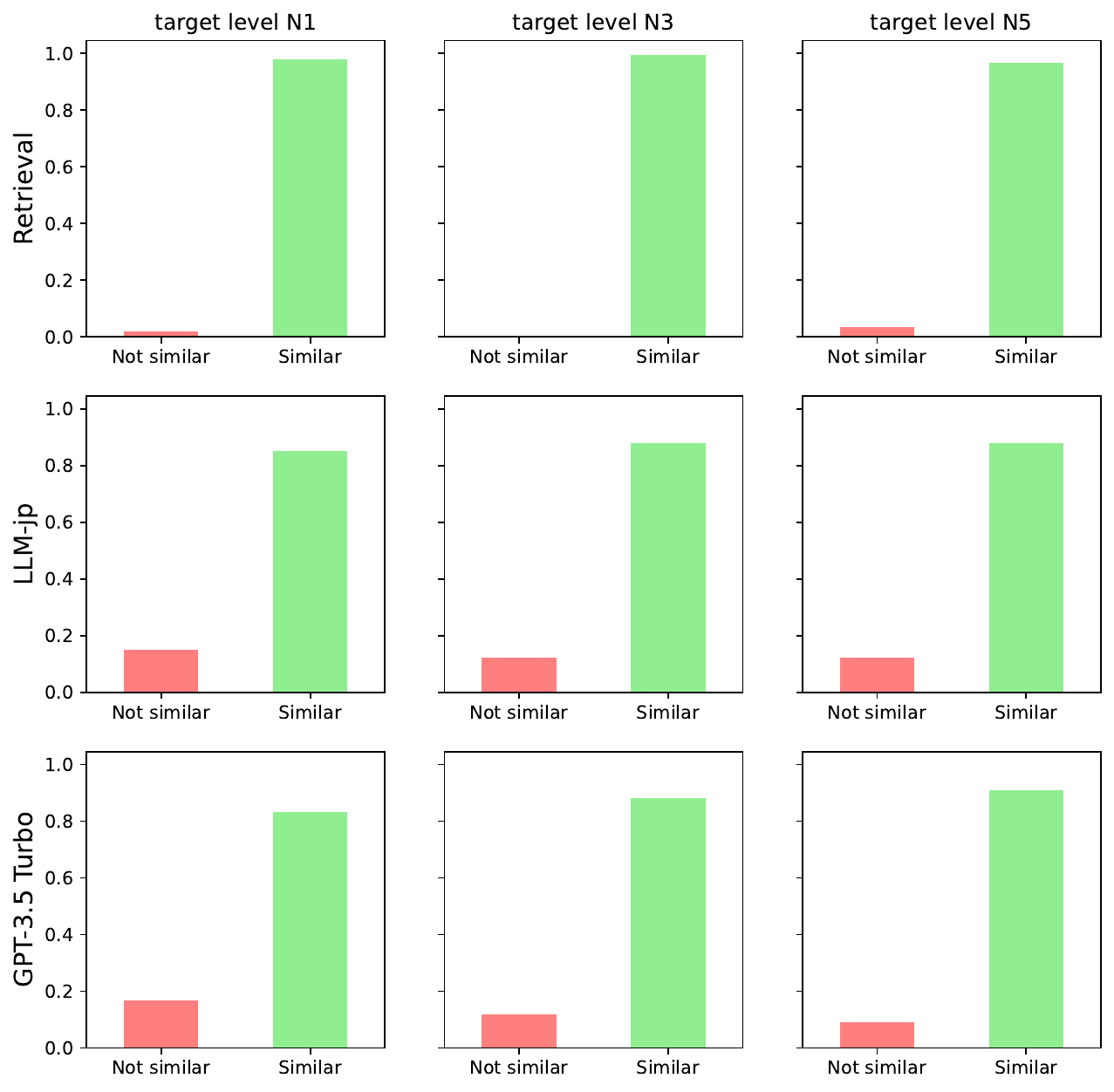}
    \caption{Ratings on sense similarity of proposed sentences.}
    \label{fig:app-sense}
\end{figure}

\begin{figure}[h!]
    \centering
    \includegraphics[max size={\textwidth}{0.42\textheight}]{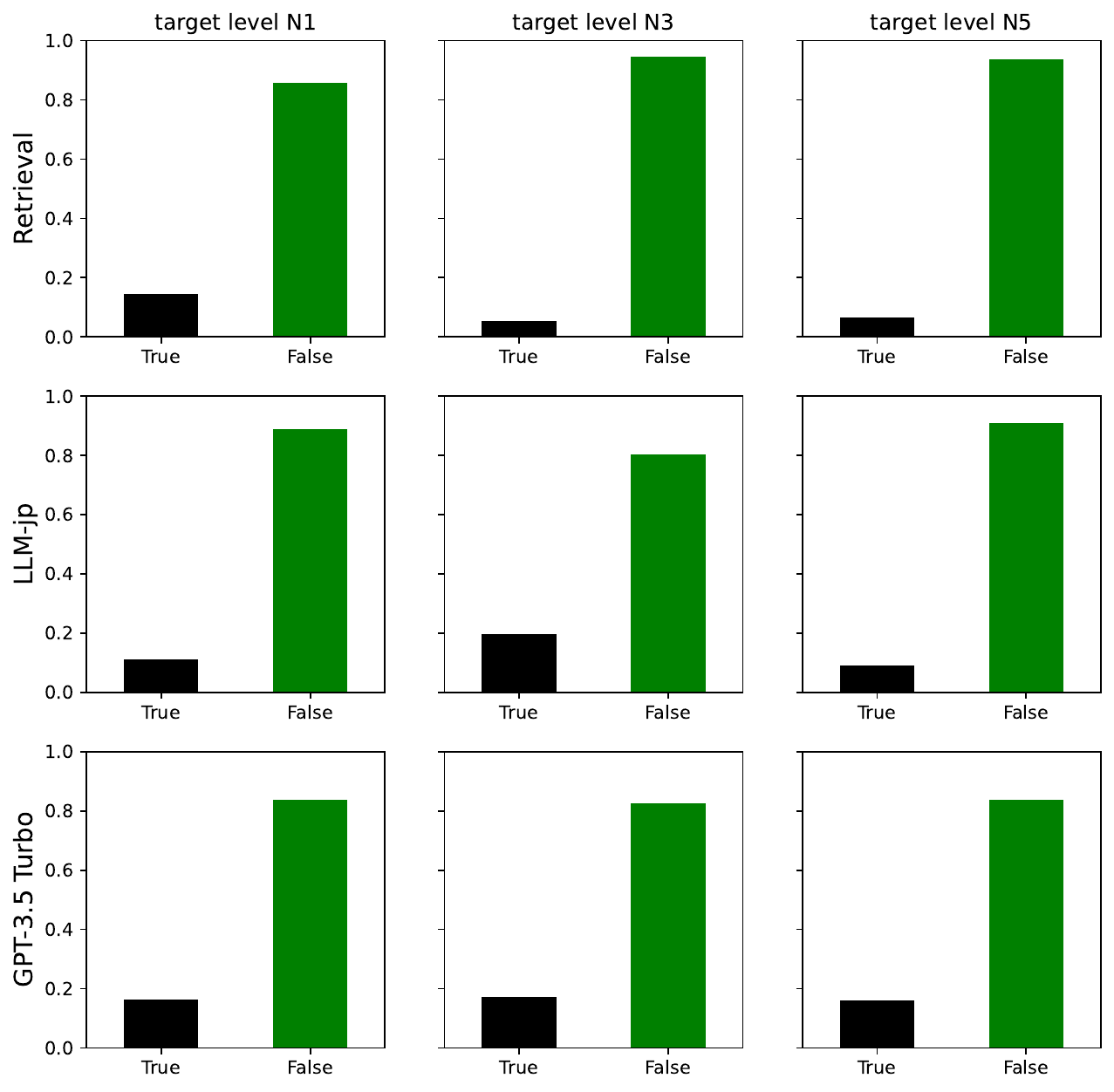}
    \caption{Proportion of rejected proposed sentences.}
    \label{fig:app-reject}
\end{figure}

\begin{figure}[ht!]
    \centering
    \includegraphics[max size={\textwidth}{0.42\textheight}]{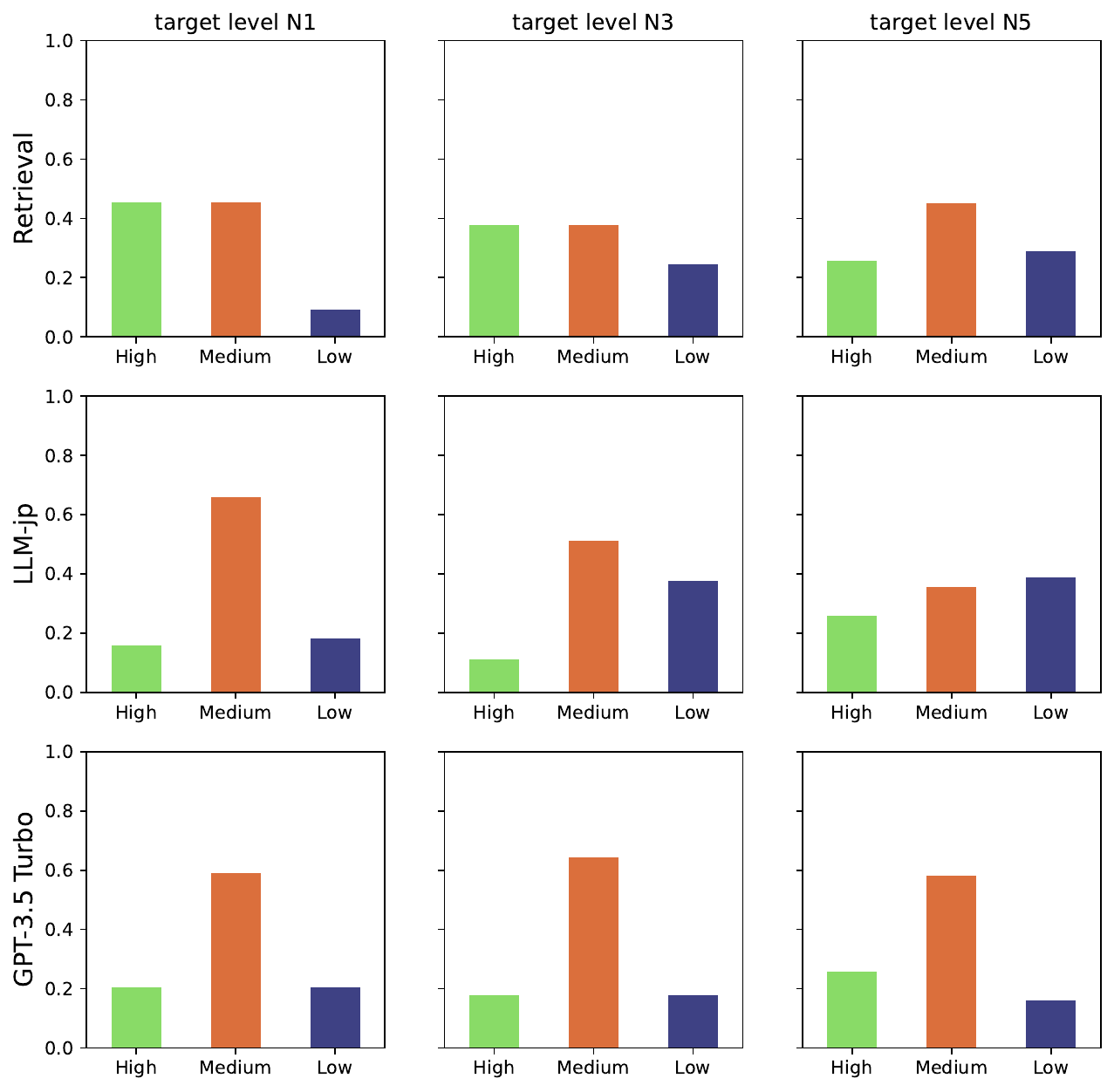}
    \caption{Ratings on syntax diversity of proposed sentences.}
    \label{fig:app-diversity}
\end{figure}

\begin{figure}[ht!]
    \centering
    \includegraphics[max size={\textwidth}{0.42\textheight}]{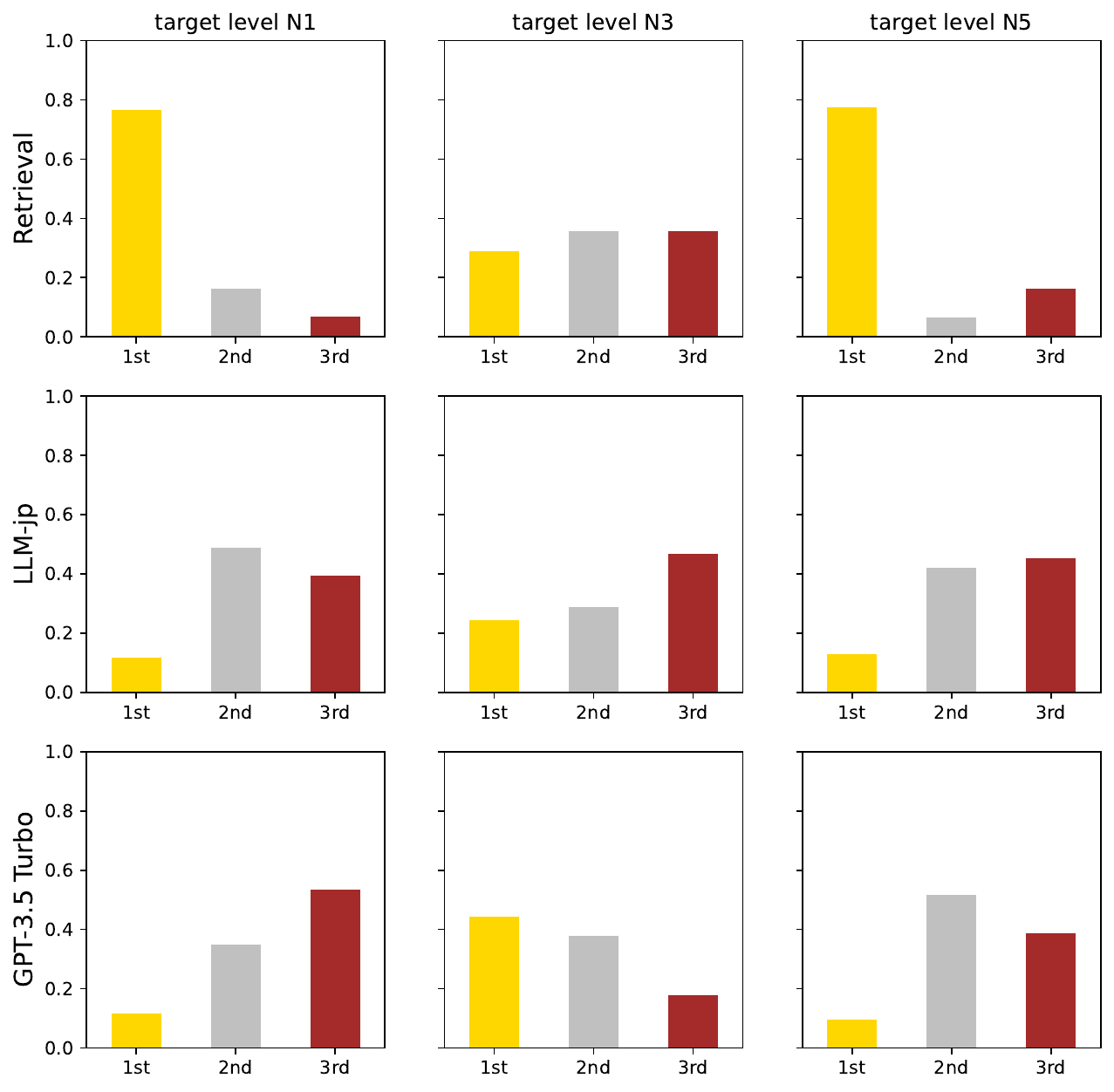}
    \caption{Rankings (first, second, third place) for each system.}
    \label{fig:ranking}
\end{figure}

% to fit in 2 pages only
% \begin{figure}[h!]
%     \centering
%     \includegraphics[width=\textwidth, height=0.45\textheight]{imgs/plots_human_all/sense_human_ann_human.pdf}
%     \caption{Ratings on sense similarity of proposed sentences.}
%     \label{fig:app-sense}
% \end{figure}

% \begin{figure}[h!]
%     \centering
%     \includegraphics[width=\textwidth, height=0.45\textheight]{imgs/plots_human_all/reject_ann_human.pdf}
%     \caption{Proportion of rejected proposed sentences.}
%     \label{fig:app-reject}
% \end{figure}

% \begin{figure}[h!]
%     \centering
%     \includegraphics[width=\textwidth, height=0.45\textheight]{imgs/plots_human_all/diversity_human_ann_human.pdf}
%     \caption{Ratings on syntax diversity of proposed sentences.}
%     \label{fig:app-diversity}
% \end{figure}

% \begin{figure}[h!]
%     \centering
%     \includegraphics[width=\textwidth, height=0.45\textheight]{imgs/plots_human_all/rank_human_ann_human.pdf}
%     \caption{Rankings (first, second, third place) for each system.}
%     \label{fig:ranking}
% \end{figure}

\end{document}